\documentclass[sn-mathphys,Numbered]{sn-jnl}

\renewenvironment{table}[1][]%
{\tableorg[#1]%
\tablebodyfont%
\renewcommand\footnotetext[2][]{{\removelastskip\vskip3pt%
\let\tablebodyfont\tablefootnotefont%
\hskip0pt\if!##1!\else{\smash{$^{##1}$}}\fi##2\par}}%
}{\endtableorg}

\usepackage{graphicx}%
\usepackage{tabularx}
\usepackage{multirow}%
\usepackage{amsmath,amssymb,amsfonts}%
\usepackage{amsthm}%
\usepackage{mathrsfs}%
\usepackage[title]{appendix}%
\usepackage{xcolor}%
\usepackage{textcomp}%
\usepackage{manyfoot}%
\usepackage{booktabs}%
\usepackage{algorithm}%
\usepackage{algorithmicx}%
\usepackage{algpseudocode}%
\usepackage{listings}%

\begin{document}

\title[Article Title]{Prediction of Arabic Legal Rulings using Large Language Models}

\author*[1]{\fnm{Adel} \sur{Ammar}}\email{aammar@psu.edu.sa}
\author[1]{\fnm{Anis} \sur{Koubaa}}\email{akoubaa@psu.edu.sa}
\author[1]{\fnm{Bilel} \sur{Benjdira}}\email{bbenjdira@psu.edu.sa}
\author[1]{\fnm{Omar} \sur{Najar}}\email{onajar@psu.edu.sa}
\author[1]{\fnm{Serry} \sur{Sibaee}}\email{218110246@psu.edu.sa}
\affil*[1]{\orgdiv{Department of Computer Science}, \orgname{Prince Sultan University}, \orgaddress{\street{Rafha Street}, \city{Riyadh}, \postcode{11586}, \country{Saudi Arabia}}}

\abstract{In the intricate field of legal studies, the analysis of court decisions is a cornerstone for the effective functioning of the judicial system. The ability to predict court outcomes helps judges during the decision-making process and equips lawyers with invaluable insights, enhancing their strategic approaches to cases. Despite its significance, the domain of Arabic court analysis remains under-explored. This paper pioneers a comprehensive predictive analysis of Arabic court decisions on a dataset of 10,813 commercial court real cases, leveraging the advanced capabilities of the current state-of-the-art large language models. Through a systematic exploration, we evaluate three prevalent foundational models (LLaMA-7b, JAIS-13b, and GPT3.5-turbo) and three training paradigms: zero-shot, one-shot, and tailored fine-tuning. Besides, we assess the benefit of summarizing and/or translating the original Arabic input texts. This leads to a spectrum of 14 model variants, for which we offer a granular performance assessment with a series of different metrics (human assessment, GPT evaluation, ROUGE, and BLEU scores). We show that all variants of LLaMA models yield limited performance, whereas GPT-3.5-based models outperform all other models by a wide margin, surpassing the average score of the dedicated Arabic-centric JAIS model by 50\%. Furthermore, we show that all scores except human evaluation are inconsistent and unreliable for assessing the performance of large language models on court decision predictions. This study paves the way for future research, bridging the gap between computational linguistics and Arabic legal analytics.}


\maketitle

\section{Introduction}
The fusion of law, artificial intelligence (AI) and natural language processing (NLP) stands as a groundbreaking frontier in contemporary research. The legal domain, with its intricate statutes, precedents, and interpretations, offers a unique challenge for computational models. Yet, the potential implications of successfully navigating this domain are profound. If legal decisions can be predicted with high precision using machine learning models, a set of invaluable insights are given to the judicial system. Such advancements could advance legal research, case preparation, and help judges and lawyers with deeper insights that they may not take in consideration during the cases' analysis. 

Predicting court decisions is challenging, especially for under-represented languages in NLP studies, such as Arabic. The inherent complexity of case description texts, combined with the nuances of the Arabic language, compounds the difficulty. Arabic, with its rich morphological structure and myriad dialects, has been a challenging landscape for NLP tasks~\cite{habash2010}. Moreover, case description texts in Arabic are characterized by their detailed rhetoric, extensive use of precedents, and domain-specific terminologies~\cite{attia2008}.

\subsection{Context}
The effectiveness of language model pre-training has been demonstrated in enhancing various tasks within the realm of natural language processing. This approach has proven successful in elevating the performance of a wide range of tasks related to processing and understanding human language \cite{dai2015semi,howard2018universal}.

In the area of AI applied to the legal domain, there have been significant advancements. The rise of machine learning has pushed in a new wave of research, with scholars exploring the potential of statistical models for legal prediction~\cite{katz2017}. The recent advancements in large language models, especially transformers, have further expanded the horizons in this domain~\cite{vaswani2017}. These models have demonstrated exceptional capabilities in a range of NLP tasks, from machine translation~\cite{sutskever2014,Afzaal_psu} to sentiment analysis~\cite{maas2011,Amjad_psu,Chaudhry_psu}, making their application to legal area an exciting avenue of exploration.

This paper embarks on an exploration of predicting Arabic court decisions using large language models (LLMs). By leveraging the latest in NLP and deep learning, we aim to test different approaches to using LLMs to maximize the predictive capability. 

\subsection{Related Works}
Language models (LMs) serve as the basis for various language technologies, but the understanding of their capabilities, limitations, and risks is still lacking. Several benchmarks were built to bridge this gap. The objective of a benchmark is to set a standard by which the performance of systems can be evaluated across a variety of tasks. Kumar et al. \cite{liang2022holistic} introduced the Holistic Evaluation of Language Models (HELM) to enhance the transparency of language models. Initially, they created a taxonomy to categorize the wide range of possible scenarios  and metrics relevant to language models. Subsequently, a comprehensive subset of scenarios and metrics was selected based on coverage and feasibility, while also identifying any gaps or underrepresentation. Finally, a multi-metric approach was adopted to evaluate language models.

The Beyond the Imitation Game benchmark (BIG-bench) was introduced by Srivastava et al. \cite{srivastava2022beyond}, featuring 204 tasks contributed by 444 authors from 132 institutions. These tasks covered diverse topics and aimed to test the limits of current language models. The performance of various model architectures, including OpenAI's GPT models and Google's dense and sparse transformers, was evaluated on BIG-bench across a wide range of model sizes. Human expert raters also participated to establish a strong baseline. The findings revealed that model performance and calibration improved with larger model sizes, but they still fell short compared to human performance. Interestingly, performance was similar across different model classes, with some advantages observed for sparse transformers. Tasks that showed gradual improvement often required extensive knowledge or memorization, while tasks with breakthrough behavior involved multiple steps or components. In settings with ambiguous context, social bias tended to increase with scale, but it could be mitigated through prompting techniques.

Elmadany et al. \cite{elmadany2022orca} presented ORCA, which is an openly accessible benchmark aimed at evaluating Arabic language comprehension. ORCA was meticulously developed to encompass various Arabic dialects and a wide range of complex comprehension tasks. It leveraged 60 distinct datasets across seven clusters of Natural Language Understanding (NLU) tasks. To assess the current advancements in Arabic NLU, ORCA was employed to conduct a thorough comparison of 18 multilingual and Arabic language models. Furthermore, a public leaderboard was provided, featuring a unified evaluation metric (ORCA score). This score represents the macro-average of the individual scores across all tasks and task clusters. 

Abdelali et al. \cite{abdelali2023benchmarking} evaluated the performance of Foundation Models (FMs) in various text and speech tasks related to Modern Standard Arabic (MSA) and Dialectal Arabic (DA), including sequence tagging and content classification, across different domains. ChatGPT (OpenAI's GPT-3.5-turbo), Whisper (OpenAI) \cite{radford2022robust}, and USM (Google) \cite{zhang2023google} were used to conduct zero-shot learning and address 33 distinct tasks using 59 publicly available datasets, resulting in 96 test setups. They found out that LLMs performed lower compared to state-of-the-art (SOTA) models, across most tasks, dialects and domains, although they achieved comparable or superior performance in a few specific tasks. The study emphasized the importance of prompt strategies and post-processing for enhancing the performance of FMs and provided in-depth insights and findings.

On another hand, the field of prompt engineering \cite{white2023prompt} has gained prominence in developing and refining inputs for language models. It provides a user-friendly and intuitive interface for human interaction with LLMs. Given the sensitivity of models to even minor changes in input, prompt engineering focuses on creating tools and techniques to identify robust prompts that yield high-performance outcomes. Various automatic optimization approaches \cite{zhou2022large,shin2020autoprompt} have been suggested to determine the optimal prompt for a particular task or a range of tasks. These methods aim to find the most suitable prompt that yields the best performance outcome.

More specifically, numerous studies have ventured into predicting court decisions across different jurisdictions. In the US, machine learning has been used to anticipate the outcomes of Supreme Court decisions~\cite{lauderdale2014}. In Europe, deep learning models have been employed to predict decisions of the European Court of Human Rights~\cite{medvedeva2019}. Concerning the Arabic legal domain, A pioneering model named AraLegal-BERT \cite{alqurishi2022aralegalbert} was proposed. It is a bidirectional encoder Transformer-based model (BERT\cite{devlin2019bert}) fine-tuned for the Arabic legal domain. The model was evaluated against three BERT variations for Arabic across three Natural Language Understanding (NLU) tasks, showcasing superior accuracy over the general and original BERT models on legal text. This work exemplifies how domain-specific customization can significantly improve language model performance in narrow domains, advancing the field's understanding of model adaptation for specialized use-cases. However, the tasks targeted in the study were specifically: legal text classification task, Named Entity Recognition Task, and Keywords Extraction Task. These tasks are different from the scope of our paper targeting prediction of Arabic Legal Rulings, which is more challenging and complex. Moreover, AraLegal-BERT is trained from scratch on specific Arabic datasets. This approach is different from the current study, where we tried first to profit from the LLMs advanced linguistic capabilities. Then, we tried to enhance the eliciting performance of these LLMs on Arabic legal ruling prediction using Zero-Shot and Few-Shot Learning. Up to our knowledge, the application of the aforementioned approach to the Arabic legal system remains a new field, with an attractive potential. This paper aims to bridge this gap by presenting a systematic investigation into predictive analysis of Arabic court decisions via an array of cutting-edge large language models tested on a dataset of real commercial cases.


\subsection{Contribution}
Given the aforementioned context and the gap identified in Arabic legal system analysis, our research offers the following novel contributions:

\begin{itemize}
    \item \textbf{Comprehensive Model Evaluation}: The research conducted a predictive analysis of Arabic court decisions by leveraging three prominent large language models: LLaMA-7b, JAIS-13b, and GPT-3.5-turbo, applied to a dataset comprising 10,813 real commercial court cases.
    
    \item \textbf{Significance of Text Preprocessing}: The study thoroughly investigated the potential benefits derived from summarizing and translating the original Arabic input texts, culminating in the creation of 14 distinct model variations.
    
    \item \textbf{Highlighting LLaMA's Limitations}: LLaMA models have been touted as almost equivalent to GPT models \cite{touvron2023llama}. Nevertheless, the findings of this paper reveal the intrinsic reduced performance of all LLaMA model variants compared to JAIS and GPT-3.5 on our dataset of Arabic court decisions.
    
    \item \textbf{Insights into Evaluation Metrics}: The paper offers a detailed evaluation of model performance using diverse metrics, namely human assessment, GPT evaluation, Rouge (1, 2, and L), and Bleu scores. Importantly, the research underscored the unreliability of all metrics, barring human assessment.
    
    \item \textbf{Bridging Research Domains}: This pivotal study bridges the gap between computational linguistics and Arabic legal analytics, laying a foundation for future scholarly endeavors in this interdisciplinary realm.
\end{itemize}




\section{Materials and Methods}

\subsection{Base Large Language Models}\label{sec:base_models}
LLaMA-7b \cite{touvron2023llama} (designed by Meta AI), JAIS-13b-chat \cite{sengupta2023jais} (MBZUAI University), and GPT-3.5-turbo \cite{brown2020language,ouyang2022training,koubaa_GPT3p5,floridi2020gpt} (OpenAI) are three recent representatives of a frontier of advancements in large language model (LLM) technology, each hailing from different origins with distinct architectural innovations. LLaMA-7b, an open-source LLM emanating from Meta AI, showcases a unique architectural approach with a range of models tailored for various applications. On the other hand, JAIS-13b-chat, with its focus on bilingual (Arabic and English) capabilities, offers a novel solution to Arabic-centric language processing tasks. GPT-3.5-turbo, a product of OpenAI, stands out for its optimization for chat-based applications, demonstrating a balance between performance and cost-effectiveness. Table~\ref{tab:LLM_models_characteristics} summarizes the main characteristics of these three models, providing a comparative glimpse into their architectural underpinnings, language and domain proficiency, training data, and use cases. Only JAIS was trained on a sizeable proportion (29\%) of Arabic texts. By contrast, Arabic language represented 0.03\% of GPT3's training dataset by word and character count, and 0.01\% by document count \cite{gpt3_languages}. Similar figures are assumed for GPT-3.5-turbo. Meta AI did not disclose the proportion of tokens per language in LLaMA models' training datasets, but the description of the sources of their pre-training datasets reveals that it is overwhelmingly in English \cite{touvron2023llama}. 

\begin{figure}[H]
  \centering
  \includegraphics[width=11cm]{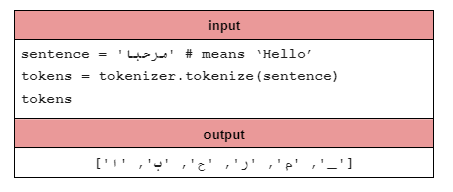}
  \caption{Over-Segmented example by LLaMA Tokenizer.}
  \label{fig:arabictokensexample}
\end{figure}

Another important element in a large language model is the tokenizer. Tokenization consists in subdiving words into sub-word tokens in order to learn vocabulary that encompasses sub-word units such as prefixes, suffixes, and root components, enabling effective handling of diverse word morphologies. Each of the three base models considered use tailored pre-trained tokenizers that are based on Byte-Pair Encoding (BPE). BPE is a data compression algorithm initially designed to reduce the size of files by replacing frequent sequences of bytes with shorter representations \cite{shibata1999byte}. In recent years, it has been adopted in NLP to tokenize text into subwords or characters in a way that strikes a balance between the flexibility of character-level representations and the efficiency of word-level representations \cite{bostrom2020byte}. Nevertheless, we noticed that most common tokenizers used in LLMs are not adapted to Arabic language, as can be seen in an example in Figure \ref{fig:arabictokensexample}. In this example, the LLaMA tokenizer segments a word into individual characters which do not have any independent meaning. The same occurs with GPT's Tiktoken tokenizer. By contrast, JAIS tokenizes the same word in this example into a single token, which conserves the meaning. 

\begin{table}[ht]
\centering
\caption{Theoretical comparison of LLaMA-7b, JAIS-13b-chat, and GPT-3.5-turbo base models.}
\label{tab:comparison}
\small
\begin{tabular}{|p{2.5cm}|p{3.5cm}|p{3.5cm}|p{3.5cm}|}
\hline
\textbf{Characteristic} & \textbf{LLaMA-7b} & \textbf{JAIS-13b-chat} & \textbf{GPT-3.5-turbo} \\
\hline
\textbf{Model Size and Architecture} & 7B Parameters, SwiGLU activation, Rotary positional embeddings. & 13B Parameters, Transformer-based decoder-only (GPT-3) architecture, SwiGLU non-linearity. & 175B parameters, GPT architecture. \\
\hline
\textbf{Language and Domain Proficiency} & Outperforms on many benchmarks including reasoning, coding, proficiency, and knowledge tests. & Bilingual (Arabic and English), State-of-the-art Arabic-centric performance. & Optimized for chat, Capable of understanding and generating natural language or code. \\
\hline
\textbf{Training Data and Open-source Availability} & Trained on 1.4 trillion tokens from publicly available datasets, overwhemingly in English. & 395B tokens (116B Arabic tokens), Pretrained with an additional 10M instruction/response pairs. & 300B tokens from various sources, overwhemingly in English. \\
\hline
\textbf{Tokenizer} & LLaMA tokenizer (BPE model based on SentencePiece \cite{kudo2018sentencepiece}). & JAIS tokenizer (BPE custom-built tokenizer that weighs both languages equally) & Tiktoken (fast optimized BPE). \\
\hline
\textbf{Use Cases and Performance} & Fine-tuned for dialogue, Optimized versions for chat (Llama-2-Chat). & Bilingual tasks, Outperforms existing open Arabic/multilingual chatbots. & Optimized for chat-based applications, Human-like responses in conversations. \\
\hline
\end{tabular}\label{tab:LLM_models_characteristics}
\end{table}

In contrast to LLaMA and GPT-3.5, the JAIS model initially refused to generate predictions concerning court decisions. This refusal reveals the type of precautionary measures incorporated into the model during its Reinforcement Learning from Human Feedback (RHLF) phase. Nevertheless, we successfully elicited predictions from the model by including an explicit instruction, stating that these are experiments that are intended solely for educational and research purposes.

We employed multiple configurations of the three aforementioned foundational models. These configurations encompass zero-shot, single-shot, and fine-tuning training paradigms. Furthermore, they are implemented on either the original Arabic dataset or on pre-processed texts that have undergone summarization and/or translation. Cumulatively, these diverse configurations result in 14 distinctive model variants. A comprehensive description of these variants is provided in Section \ref{sec:LLM-models}.

\subsection{Fine-tuning using LLM-Adapters}

Engaging in complete finetuning has the potential to result in catastrophic forgetting, given that it involves altering all parameters within the model. In contrast, Parameter Efficient Fine-Tuning (PEFT), by exclusively modifying a limited subset of parameters, as opposed to full-parameter fine tuning, demonstrates greater resilience against the detrimental impacts of catastrophic forgetting \cite{pu2023empirical}. In this context, LLM adapters offer a simple and efficient approach to PEFT in large language models \cite{hu2023llm}. LoRA (Low-Rank Adaptation) is a method that can significantly reduce the number of trainable parameters required for fine-tuning large language models. It is a type of LLM adapters that is integrated into the LLM-Adapters framework and supports fine-tuning of LLaMA models among others \cite{hu2023llm}. As LoRA has a significant motivations of successfully lowering the amount of trainable factors without sacrificing performance, applying it to the LLaMA model aims to achieve high performance while minimizing computational costs. 

With this approach, LoRA follows a strategy that reduces the number of parameters to be trained during fine-tuning by freezing all of the original model parameters and then inserting a pair of rank decomposition matrices alongside the original weights. Additionally, LoRA utilizes the adapter method in such a way of adding a subset of parameters, enabling a few low-intrinsic adapters in parallel with the attention module without increasing inference latency. In this work, we carry out the fine-tuning of the LLaMA-7b base model using LoRA approach on Arabic texts, following the implementation of \cite{github_llm_adapters}. In fact, LoRA's design allows for more flexibility in adding adapters \cite{hu2021lora}, making it efficient for scaling up to large language models for improved performance on custom datasets and tasks. Nevertheless, we did not manage to fine-tune the larger JAIS-13b and GPT-3.5-turbo base models due to resource constraints.

\begin{figure}[H]
\centering
\includegraphics[width=10cm]{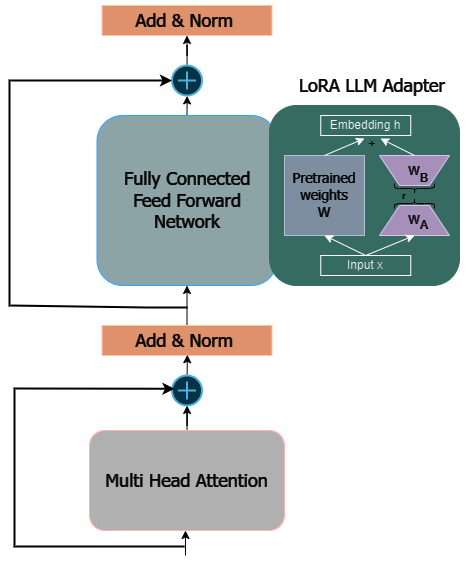}
\caption{Operational schematic of LoRa adapters within the transformer.\label{fig:diagram_1_Lora_adapter}}
\end{figure}

Figure \ref{fig:diagram_1_Lora_adapter} illustrates the integrated mechanism of the LoRa adapter within the LLM module of the transformer, highlighting the modified forward pass in the network, and the weight adjustment mechanism. The LoRa method enhances the fine-tuning of Large Language Models (LLMs) by decomposing the weight update matrix into a lower-rank representation instead of updating the original weight matrix directly, leading to fewer parameters during adaptation. This results in faster training and potentially reduced computational needs without losing vital information. In conventional fine-tuning, weight changes are computed via backpropagation based on the loss gradient. LoRa, instead, decomposes these changes into two smaller, lower-dimensional matrices. Then, it trains these smaller matrices, enabling effective representation in a lower-dimensional space and reducing the parameter space.

In the LoRA method, the decomposition of the weight update matrix \( \Delta W \) into two matrices \( W_A \) and \( W_B \) is given by:
\begin{equation} \label{eq:decomp}
\Delta W = W_A W_B.
\end{equation}

Assuming \( W_A \) and \( W_B \) are of dimensions \( m \times r \) and \( r \times n \) respectively, where \( r \) is the rank, and \( m \) and \( n \) are the dimensions of the original matrix \( \Delta W \), the total number of parameters to be learned reduces from \( m \times n \) to \( m \times r + r \times n \).

Further, if \( X \) is the input to a layer and \( Y \) is the output, the modified forward pass in LoRA can be represented as:
\begin{equation} \label{eq:lora_forward}
Y = (W + W_A W_B)X + b,
\end{equation}
where \( b \) is the bias vector.

The error \( E \) in approximation can also be analyzed. It's given by:
\begin{equation} \label{eq:error}
E = ||\Delta W - W_A W_B||_F,
\end{equation}
where \( ||.||_F \) denotes the Frobenius norm.
This mathematical formulation elucidates the reduction in computational complexity and the preservation of essential information for task adaptation achieved by LoRA. This method preserves the essential information required for task adaptation while reducing the computational burden, showcasing a trade-off between model complexity and adaptation capacity.

 The implementation of LoRa is relatively straightforward, as seen in Figure \ref{fig:diagram_1_Lora_adapter}. A modified forward pass in the network is applied, adjusting the magnitude of weight updates to balance pre-trained knowledge with new task-specific adaptation.

\subsection{Dataset}

We retrieved the Saudi Ministry of Justice dataset (SMOJ) through a web scraping from the Saudi Justice Portal (SJP) website \cite{SJP}, focusing on the category of commercial courts, which contains a series of court decisions about financial and commercial disputes, all in Arabic language. To facilitate the data retrieval, we used Selenium Python library \cite{selenium}, which enables programmatic interactions with web pages, essentially simulating user actions to access and gather data.

The data collection process for SMOJ was structured and systematic, starting with the iteration through a range of page numbers. This range spans from page 1 to 60,000, a scope determined based on the expected volume of data available on the website. Before any data extraction occurs, each page's availability is verified by checking for the presence of the text 'Page not found.' This precautionary measure ensures that only existing pages are processed, minimizing potential errors and preventing unnecessary resource consumption.

Once page availability is confirmed, the data extraction process is initiated using Beautiful Soup Python package \cite{bs4}, which is tailored for HTML and XML parsing, and is employed to dissect the HTML structure of the web pages. This allows for the extraction of specific elements, focusing on critical legal information contained within the SJP website. The data extraction process focuses on three primary categories: case description, justification, and court decision. We used the case description as input (prompt) to the LLM models, and the court decision as output (completion). There is no strictly pre-defined format or ordering for the case description and court decision. We decided to ignore the justification field and not include it in the input, seeing that it often unveils the inclination of the court decision. 

After removing duplicates and excessively long cases (more than 4096 tokens), we randomly subdivided the SMOJ dataset into a training dataset containing 10,713 cases and a testing dataset containing 100 cases. We opted for a reduced testing dataset to be able to manually evaluate the outputs of each LLM model. In fact, we will shown in section \ref{sec:Results} that all other automatic evaluations were unreliable and inconsistent.

\begin{figure}[H]
\includegraphics[width=15 cm]{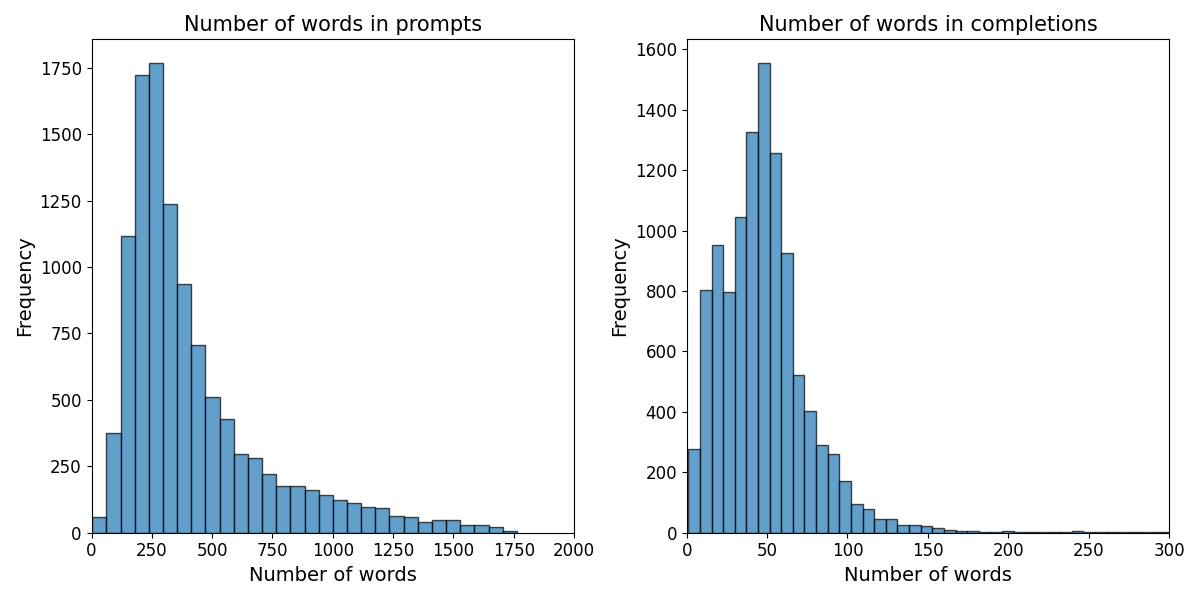}
\caption{Histogram of the number of words in the prompts (case descriptions) and completions (court decisions) in the SMOJ training dataset.\label{fig:hist_nb_words}}
\end{figure}

Figure \ref{fig:hist_nb_words} depicts the histogram of the number of words in the prompts (case descriptions) and completions (court decisions) in the SMOJ training dataset. The total number of words in the training set is 5M, and the average number of words in the prompts and completions is 422 and 52, respectively. The large size of the prompts is a real challenge, which motivated us to test LLM models on summarized prompts, as will be detailed in section \ref{sec:LLM-models}.

\subsection{LLM model variants}\label{sec:LLM-models}


\begin{figure}[h]
\includegraphics[width=16 cm]{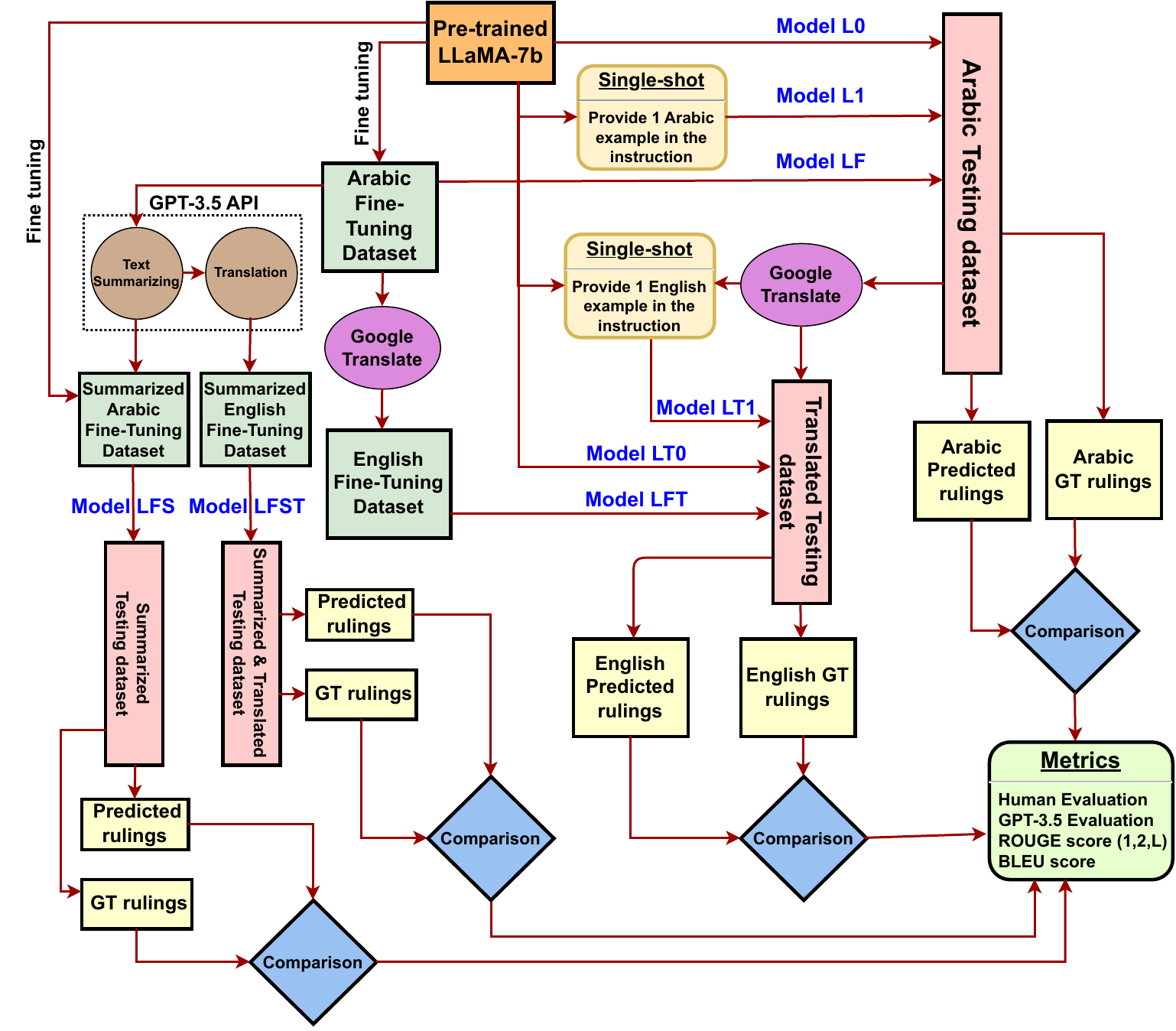}
\caption{Diagram of the main steps for evaluating the 8 LLaMA variant models on the SMOJ dataset.\label{fig:diagram_8_LLaMA_models}}
\end{figure}   

For each of the three base pre-trained models described in section \ref{sec:base_models} (LLaMA-7b, GPT-3.5-turbo, and JAIS-13b-chat), we implemented different variants. Figure \ref{fig:diagram_8_LLaMA_models} illustrates the main steps for evaluating the 8 LLaMA variant models on the SMOJ dataset. The base pre-trained LLaMA-7b model is the core of all these models. They differ by the inclusion or not of single-shot or fine-tuning learning, and the addition or not of summarizing and/or translation steps:

\begin{itemize}
    \item \textbf{Model L0} is a zero-shot model. It is the mere application of the base pre-trained LLaMA-7b model on each prompt of the Arabic testing dataset without any pre-processing or learning steps.
    \item \textbf{Model L1} is a single-shot variant, where a single prompt/completion pair from the original Arabic training dataset is provided in the instruction, to act as an example to follow.
    \item \textbf{Model LT0} is a zero-shot model applied to an English testing dataset. This dataset was obtained using Google Translate API through Python \textit{translators} package \cite{translators_pypi}. The translation of the original Arabic dataset into English can be beneficial to enhance the prediction for LLaMA and GPT-3.5 models, since they are overwhelmingly pre-trained on English texts, as explained in section \ref{sec:base_models}. The assessment of the usefulness of this pre-processing step will be discussed in section \ref{sec:quant_eval}.
    \item \textbf{Model LT1} is a single-shot model applied to the translated English testing dataset. It includes in the instruction a single translated prompt/completion pair from the training dataset.
    \item \textbf{Model LF} is obtained by fine-tuning the base model on the original Arabic training dataset for 200 epochs.
    \item \textbf{Model LFT} is obtained by fine-tuning the base model on the translated English training dataset for 200 epochs.
    \item \textbf{Model LFS} is obtained by fine-tuning the base model on a subset of the Arabic training dataset after summarizing the prompts through GPT-3.5-turbo API. We selected only a subset of 1000 prompt and completion pairs due to budget limitations, since requests to the GPT API are costly.
    \item \textbf{Model LFST} is obtained by fine-tuning the base model on a subset of 1000 prompt and completion pairs from the Arabic training dataset after summarizing and translating them through GPT-3.5-turbo API. 
    
 \end{itemize}

 Table \ref{tab:Instructions_for_summarization_and_translation} showcases the instructions fed to GPT-3.5-turbo API to summarize and/or translate the original SMOJ dataset, for fine-tuning the LFS and LFST models.

\begin{table}[H]
 \centering
 \caption{Instructions used to summarize and translate the SMOJ dataset through GPT-3.5-turbo API.\label{tab:Instructions_for_summarization_and_translation}}
\includegraphics[width=12 cm]{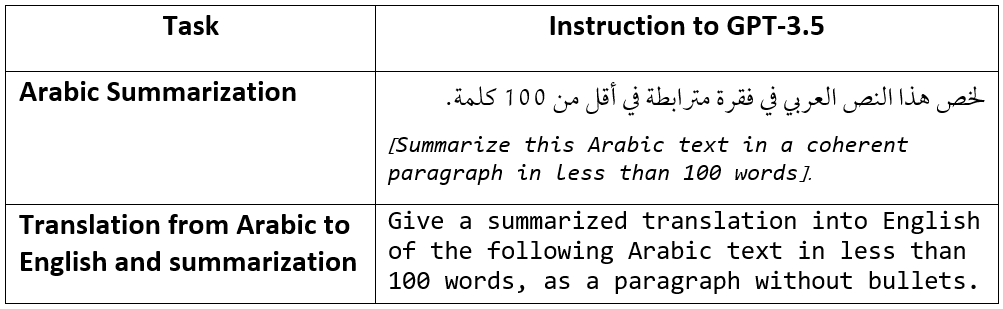}
\end{table}  
 
Similarly, models G0, G1, GT0, and GT1 are obtained in the same way as L0, L1, LT0, and LT1, respectively, but using GPT-3.5-turbo as a base model, instead of LLaMA-7b. Likewise, J0 and J1 are zero-shot and single-shot variants of JAIS-13b-chat model. 

Due to limited resources, we could not fine-tune the GPT and JAIS models in the same way that we did for the LLaMA models. Besides, it is pointless to apply JAIS on translated text, since it was pre-trained with special focus on Arabic language. Furthermore, multi-shot variants were not examined in our research. This is due to the extensive input size present in our dataset and the inherent limited context length associated with the models (4096 tokens).


For each model, we employ a suite of metrics, as detailed in section \ref{sec:metrics}, to evaluate their performance by comparing the predicted rulings to the suitable version of the ground-truth (GT) rulings from the test dataset. Specifically:

\begin{itemize}
    \item Models L0, L1, LF, J0, J1, G0, and G1 are evaluated against the original Arabic version of the GT rulings.
    \item Model LFS is gauged against the summarized Arabic version.
    \item Models LT1, LT0, LFT, GT0, and GT1 are assessed based on the translated form of the GT rulings.
    \item Model LFST is measured against the GT rulings that have been both summarized and  translated.
\end{itemize}

\subsection{Metrics} \label{sec:metrics}
The following metrics were applied to evaluate each of the LLM models described in section \ref{sec:LLM-models}:
\begin{itemize}
    \item \textbf{Human score:} A human evaluator was tasked with assessing the accuracy of the predicted rulings generated by each model in relation to the ground-truth rulings of the test dataset. This evaluation was conducted on a scale ranging from 0 to 5. A score of 0 indicated that the predicted ruling was either nonsensical or wholly incorrect, while a score of 5 signified a flawless prediction, mirroring the decisions encapsulated in the ground-truth ruling, regardless of the actual wording. To ensure a uniform evaluation standard and minimize variability in scoring, all model outputs were reviewed by the same evaluator.
    \item \textbf{GPT score:} We used GPT-3.5-turbo API to automatically and systematically compare all the predicted rulings generated by each model to the ground-truth rulings of the test dataset. To guide this assessment, we provided the GPT model with the following  instruction: \textit{"Compare the following two court decisions (predicted: 'Decision (predicted)' and ground-truth: 'Decision (GT)') and assign a score from 0 to 5 to the predicted decision. 0 means: Non sense. 5: means perfect answer. Format the response as: Score; Justification. For example: 0; Non sense."} 
    \item \textbf{BLEU score:} BLEU \cite{papineni2002bleu, chen2014systematic}, an acronym for Bilingual Evaluation Understudy, was designed as a metric for assessing the quality of machine-translated text between two natural languages. The BLEU score is computed using a weighted geometric mean of modified n-gram precisions. This is further adjusted by the brevity penalty, which diminishes the score if the machine translation is notably shorter than the reference translation. The utilization of the weighted geometric mean ensures a preference for translations that consistently perform well across different n-gram precision levels. More specifically, the BLEU score is given by:
\[ \text{BLEU} = BP \times \exp \left( \sum_{n=1}^{N} w_n \log p_n \right) \]

Where:
\begin{itemize}
    \item \( p_n \) is the n-gram precision.
    \item \( w_n \) are the weights for each precision (typically \( w_1 = w_2 = w_3 = w_4 = 0.25 \) for BLEU-4).
    \item \( BP \) is the brevity penalty:
    \[ BP = 
    \begin{cases} 
      1 & \text{if } c > r \\
      \exp(1 - \frac{r}{c}) & \text{if } c \leq r 
    \end{cases}
    \]
    with \( c \) as the predicted output length and \( r \) as the ground-truth length.
\end{itemize}
    
    We calculated the BLEU metric using the \textit{nltk.translate.bleu\_score} python module \cite{bleu_nltk}.
      \item \textbf{ROUGE score}: ROUGE \cite{lin2004rouge} is an acronym for Recall-Oriented Understudy for Gisting Evaluation. Its primary purpose is to evaluate the performance of automatic summarization tools and machine translation systems within the realm of natural language processing (NLP). The fundamental idea behind ROUGE is to juxtapose an algorithmically generated summary or translation with one or multiple human-crafted reference summaries or translations. This comparison helps to determine how well the machine-generated output aligns with the human standard. We apply it here to the comparison between predicted and GT rulings in the SMOJ testing dataset. We specifically used three variants of the ROUGE score:
    \begin{itemize}
        \item \textbf{ROUGE-1}: This metric gauges the overlap of unigrams (individual words) between the predicted output and the GT ruling. By examining the matching single words between both texts, ROUGE-1 provides insights into the basic lexical similarity.
        
        \item \textbf{ROUGE-2}: Stepping beyond individual words, ROUGE-2 considers bigrams (pairs of adjacent words). By comparing the overlap of these word pairs between the predicted and GT ouputs, ROUGE-2 offers a deeper understanding of the phrasal and structural alignment.
        The general formula for the ROUGE-N score is:
        \[
\text{ROUGE-N} = \frac{\sum_{s \in \text{GT}} \sum_{\text{N-gram} \in s} \text{Count}_{\text{match}}(\text{N-gram})}{\sum_{s \in \text{GT}} \sum_{\text{N-gram} \in s} \text{Count}(\text{N-gram})}
\]
Where:
\begin{itemize}
    \item \(\text{Count}_{\text{match}}(\text{N-gram})\) is the maximum number of times an N-gram is found in both the predicted and GT ouputs.
    \item \(\text{Count}(\text{N-gram})\) is the count of the N-gram in the GT ouput.
\end{itemize}
        
        \item \textbf{ROUGE-L}: This metric employs the concept of the Longest Common Subsequence (LCS). LCS is the maximum sequence of tokens that appear in both the machine-produced and reference texts. This metric offers a more holistic perspective on similarity as it naturally considers sentence-level structures and automatically identifies the co-occurring n-gram sequences.
    \end{itemize}
    For each of these three metrics, we compute the precision (P), recall (R), and F1-score (F). We calculated the ROUGE metrics using the \textit{rouge 1.0.1} Python library \cite{rouge_pypi}.
\end{itemize}
While the BLEU and ROUGE metrics were primarily conceived for tasks related to translation and summarization, they can potentially serve as indicative tools for evaluating the alignment between predicted and GT rulings in the SMOJ dataset. We will assess the correctness of this hypothesis in section \ref{sec:Results}.






\section{Results}\label{sec:Results}
Within this section, we undertake a comprehensive evaluation of the implemented models, encompassing both qualitative and quantitative assessments. In subsection \ref{sec:qual_eval}, we present a qualitative comparison between human and GPT scores on a small sample, exemplifying scenarios where predictions align with or deviate from expectations. We also discuss the challenges and nuances of employing GPT-3.5 as an evaluation metric. In subsection \ref{sec:quant_eval}, we delve into the performance evaluation of the 14 models using diverse metrics, shedding light on the impact of zero-shot, single-shot, and fine-tuning approaches, as well as the prompt summarization and/or translation pre-processing steps. We further discuss the reliability of GPT, BLEU, and ROUGE scores. This holistic evaluation provides insights into the strengths and limitations of LLMs for the prediction of court decisions.







\subsection{Qualitative evaluation}\label{sec:qual_eval}

Table \ref{tab:sample_evaluation_scores} provides a qualitative comparison between human and GPT scores on a small sample of predicted and GT rulings from the testing dataset. This sample is representative of most of the encountered cases. The first row shows an example in Arabic. The predicted output contains a correct decision briefly expressed with implicit reference to the amount mentioned in the input (case description), while the GT ruling explicitly mentions the names of the plaintiff and defendant and the amount of money that the latter should pay to the former. Because of the difference in formulation, the GPT API gave the prediction a score of only 2/5. Whereas, the human evaluator took into account the semantic matching and assigned a higher score of 4/5.

In the second example, the LLM model issues a perfect ruling matching the same amount to be paid by the plaintiff as in the GT decision. Even though the identities of the plaintiff and defendant are not explicitly mentioned, this is not important, since they are already mentioned in the case description. In this case, both the human evaluator and GPT-3.5 assign a perfect score of 5/5.

In the third example, the predicted output is a series of nonsensical words and symbols. This happens often with LLaMA models, especially when the input size is large. As expected, the human score in this case is 0. However, GPT-3.5 oddly assigns a score of 2/5 to this prediction. This example also reveals the poor Google translation in the GT output, especially for the last sentence where the Arabic word Al-h\={a}d\=\i~('guide') was mistaken for its homonym: 'pacific'. Such translation shortcomings can affect the quality of LLM training. 

The fourth prediction example in Table \ref{tab:sample_evaluation_scores} is similar, in terms of meaningless predicted output and poor GT translation, but in this case, both the human and GPT scores are rightly equal to 0.

In the fifth and last example, the LLM model just rehashed the instruction and part of the input that was fed to it, without adding any prediction. This also often happens with LLaMA models. As expected, the human score in this case is 0. However, GPT-3.5 surprisingly assigns a score of 4/5 to this prediction, which showcases its unreliability as en evaluation metric. This will be discussed more precisely using quantitative analysis in the next section.

 \begin{table}[H]
 \centering
 \caption{Human and GPT assigned scores on sample predicted and GT rulings from the testing dataset.\label{tab:sample_evaluation_scores}}
\includegraphics[width=14 cm]{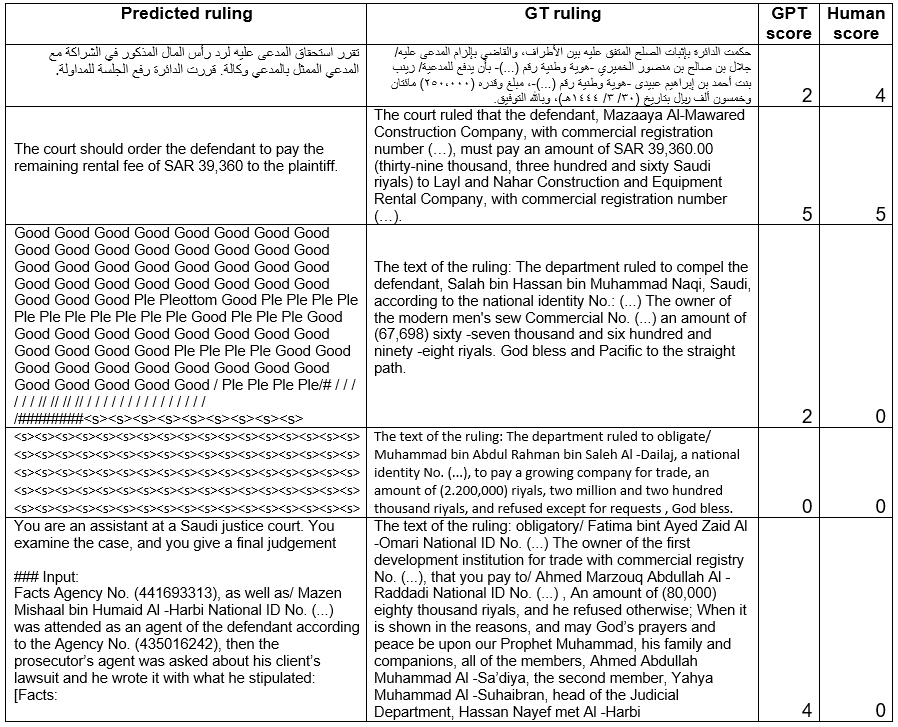}
\end{table}

\subsection{Quantitative evaluation}\label{sec:quant_eval}

Table \ref{tab:LLaMA-results} presents the performance evaluation of the eight LLaMA-7b variant models on the test datasets, measured using different metrics. The LFST model, which underwent both summarization and translation processes using GPT-3.5-turbo, consistently delivered the top performance across nearly all evaluation metrics. In stark contrast, the other LLaMA variants displayed significantly subpar results. Their human evaluation scores were under 0.5 out of 5, and their GPT scores did not surpass 2 out of 5. Furthermore, both the Rouge and Bleu scores for all models were notably low. This underperformance was particularly pronounced for Arabic language models. L1 receives a score of 0 for all metrics except GPT score. This performance degradation compared to L0 can be explained by the increase of the input size due to adding an example of prompt/completion in the instruction, which often makes the total input size exceed the model's maximum context length.

\begin{figure}[h]
\includegraphics[width=15 cm]{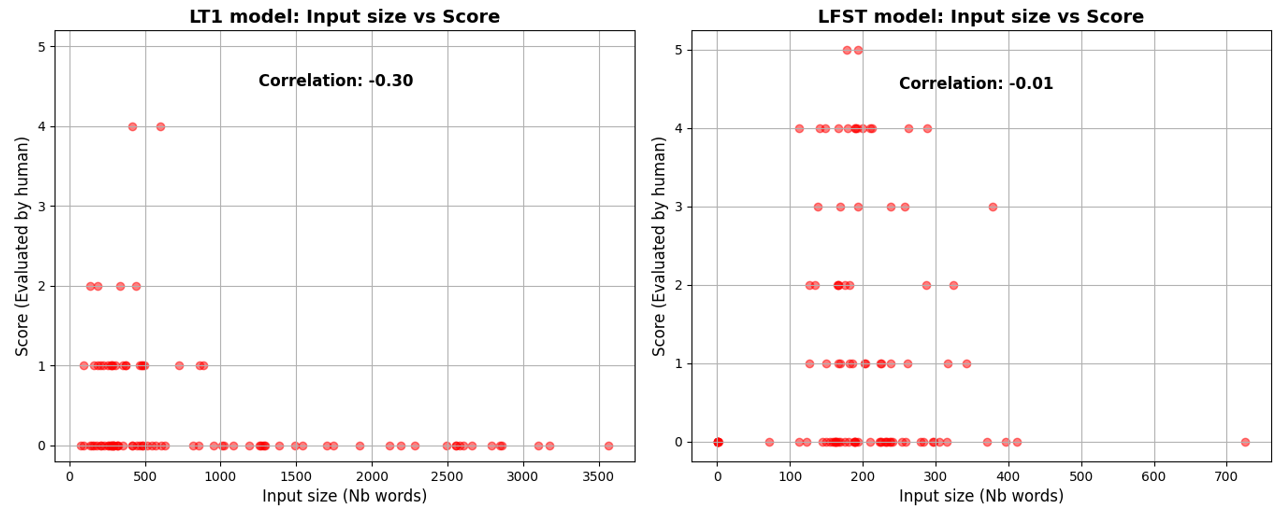}
\caption{Scatter-plot between the input size (in terms of number of words) and the human score obtained, for LT1 (left) and LFST (right) models.\label{fig:scatter-nb_words-vs-score}}
\end{figure}

The primary reason for summarizing prompts in the LFS and LFST models stems from our observation that the LLaMA models frequently produce low-quality responses to longer prompts. This observation finds some validation in Figure \ref{fig:scatter-nb_words-vs-score}, which showcases scatter plots correlating input size (measured by word count) with human evaluation scores for the LT1 and LFST models. Notably, for the LT1 model, prompts exceeding 1000 words invariably receive a score of zero. However, the overall correlation remains relatively weak, at -0.3. In contrast, upon summarizing the prompts for the LFST model, the correlation between input size and evaluation score vanishes. This suggests that the modified LLaMA model can handle moderately sized inputs in an equal manner.

\begin{table}
\centering
\caption{Results of the evaluation of the 8 LLaMA-7b variant models on the testing datasets, using various metrics.}
\includegraphics[width=15 cm]{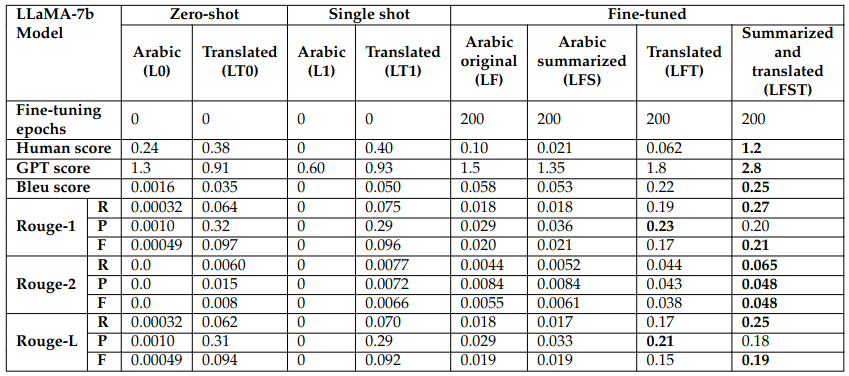}
\label{tab:LLaMA-results}
\end{table}

\unskip

Table \ref{tab:GPT-results} presents the outcomes of applying the same metrics to the four GPT-3.5-turbo variant models. Both G0 and GT1 demonstrate closely aligned performance when evaluated using human scores. This suggests that the integration of translation and single-shot training did not significantly enhance performance for GPT-based models. However, when we consider the Bleu and Rouge scores, the translated models, GT0 and GT1, consistently outperform their counterparts. Interestingly, there is a noticeable discrepancy between the GPT score and the human judgment. A more detailed examination of specific prediction instances confirms that the GPT score can be unreliable in several scenarios.

\begin{table}
\centering
\caption{Results of the evaluation of the four GPT3.5-turbo variant models on the testing datasets, using various metrics.}
\includegraphics[width=11 cm]{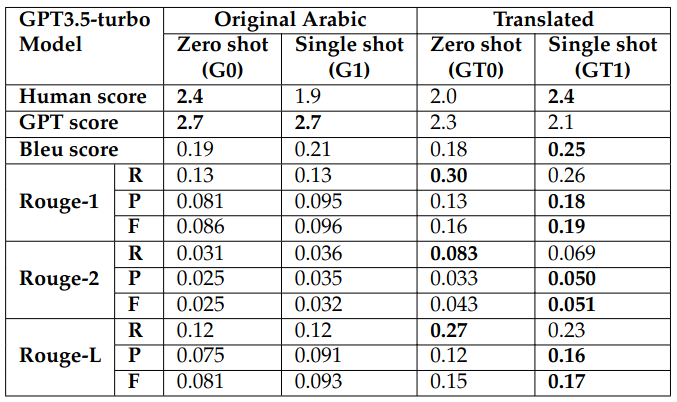}
\label{tab:GPT-results}
\end{table}

Table \ref{tab:JAIS-results} shows the performance of the two JAIS-13b-chat models. Only Arabic-based models were tested in this scenario since the JAIS base model is specifically tailored for Arabic language. We observe a slight improvement when moving form zero-shot (J0) to single-shot (J1) according to all metrics, except for the GPT score. This further highlights the unreliability of the GPT score for this task. Even though JAIS was pre-trained with special focus on Arabic language, it falls short in comparison with all GPT-based models (Table \ref{tab:GPT-results}). This confirms the superiority of GPT-based models for a wide range of tasks even for under-represented languages in its learning dataset, such as Arabic.

\begin{figure}[H]
\includegraphics[width=15 cm]{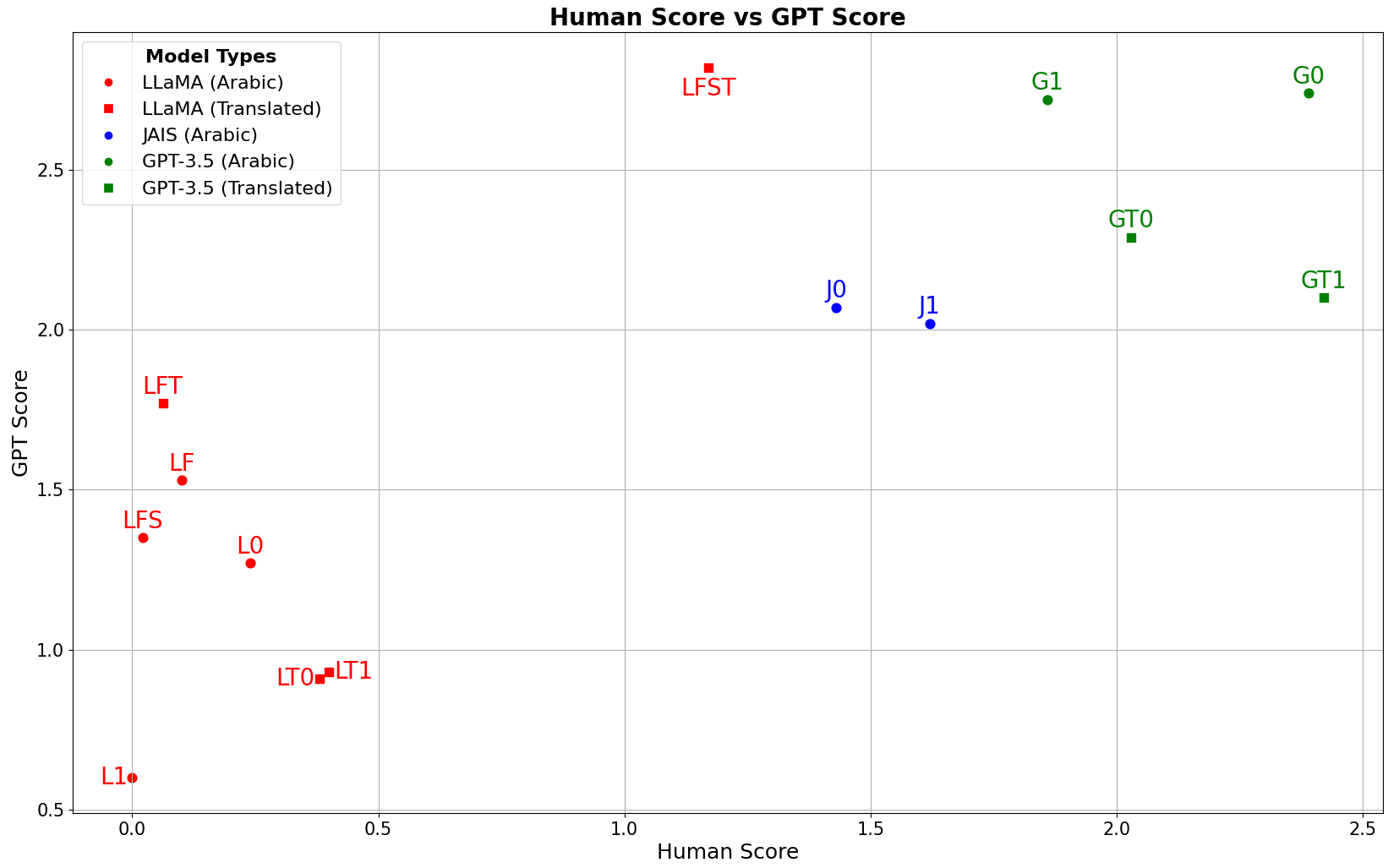}
\caption{Human score versus GPT score for each tested Large Language Model. Arabic-based models are represented as circles, while English-based models are represented as squares, with a different color code for each base model (LLaMA, JAIS, GPT-3.5).\label{fig:models_human_vs_GPT_scores}}
\end{figure}

Figure \ref{fig:models_human_vs_GPT_scores} maps out the 14 implemented models in the (Human score, GPT score) space. This visualization underscores the dominance of the GPT-based models and the underperformance of the LLaMA-based counterparts. Among the LLaMA models, only the LFST variant comes close to the performance of JAIS and GPT models in terms of human evaluation. Notably, LFST is not a pure LLaMA model as it leverages the summarizing and translation capabilities of GPT-3.5. On another hand, while JAIS models outpace LLaMA models, they lag behind the GPT models. A striking feature of Figure \ref{fig:models_human_vs_GPT_scores} is the evident discrepancy between GPT and human scores. For instance, despite LFST achieving the highest GPT score across all models, it secures a merely moderate human score. In a similar vein, LFT showcases a higher GPT score than both LT0 and LT1, even though the latter pair surpass it in human evaluations. This incongruence is especially pronounced in English-based models.

\begin{table}[H]
\centering
\caption{Results of the evaluation of the two JAIS variant models on the testing datasets, using various metrics.}
\includegraphics[width=7 cm]{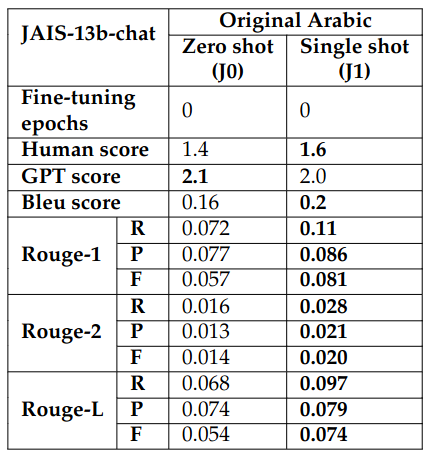}
\label{tab:JAIS-results}
\end{table}

\begin{figure}[H]
\includegraphics[width=17.5 cm]{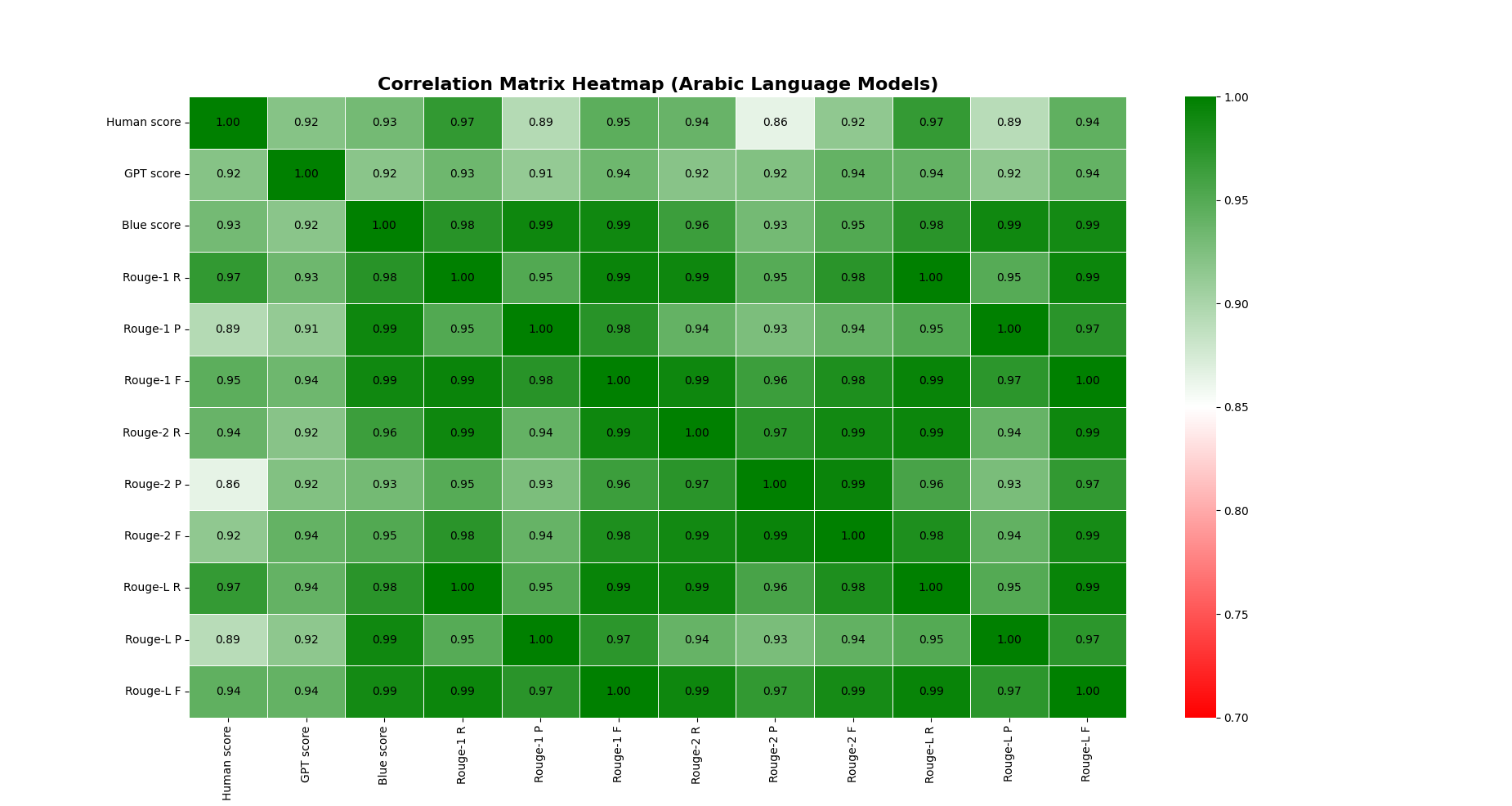}
\includegraphics[width=17.5 cm]{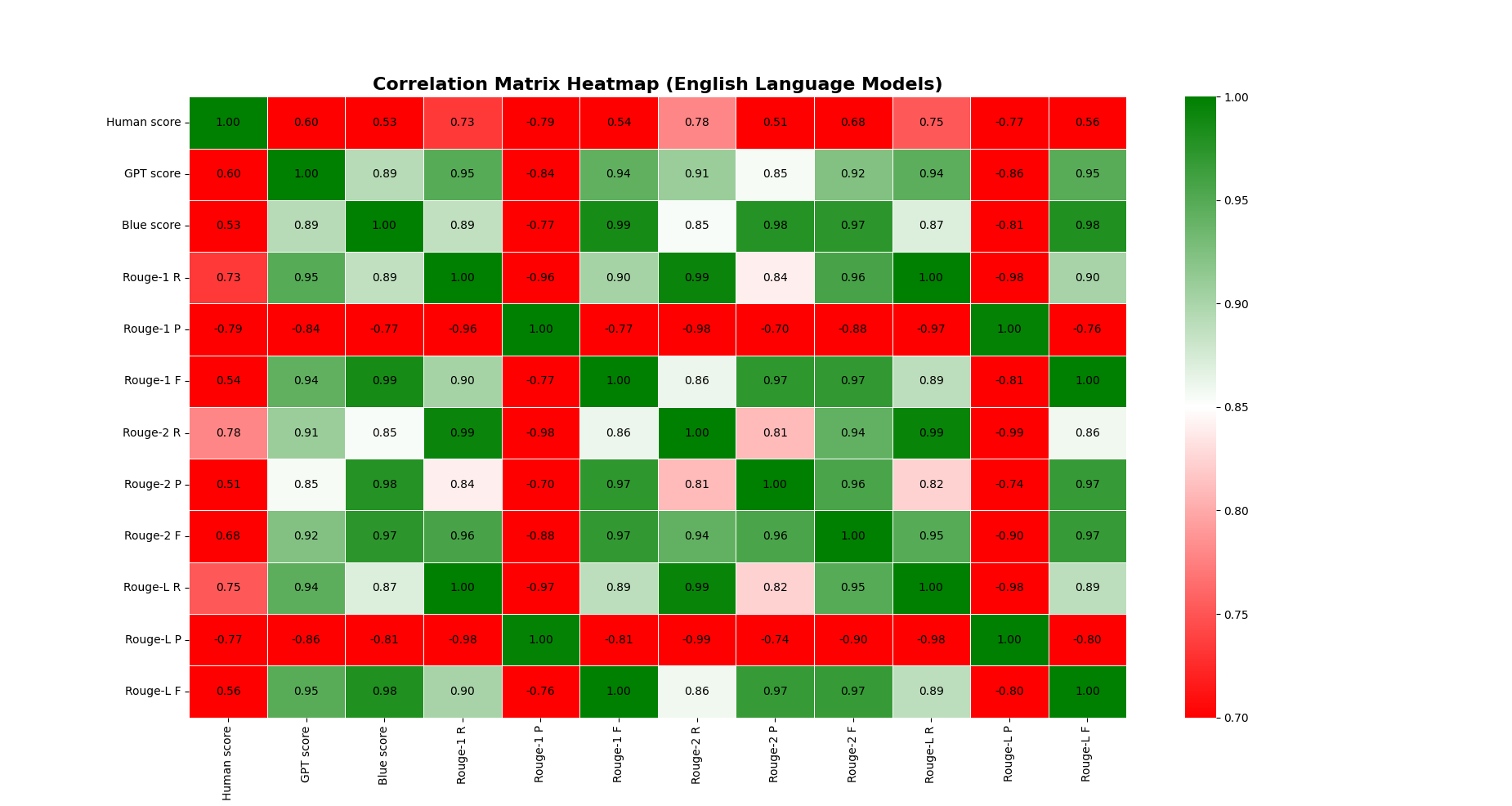}
\caption{Heatmap of the correlation coefficient between the values of the metrics used for evaluating Arabic (top) and translated (bottom) language models.\label{fig:heatmap_correlation}}
\end{figure}

This observation is further confirmed in Figure \ref{fig:heatmap_correlation} where we notice that the correlation between the human score and GPT score is much higher for Arabic-based models (0.92) than for English-based models (0.60). A plausible explanation for this divergence is that the process of translating from Arabic to English may introduce errors, omitting or misrepresenting key details, which makes the score evaluation by GPT-3.5 more challenging. This observation extends to the Bleu and Rouge scores, which consistently display a lower alignment with human scores for English models. Most notably, Rouge-1 precision and Rouge-L precision exhibit negative correlations with human scores, standing at -0.79 and -0.77 respectively. All these results suggest that the GPT, Bleu, and Rouge scores are unreliable for performance evaluation in the considered task.

\section{Discussion}
The results presented in section \ref{sec:Results} provide several important insights in the context of legal ruling prediction using large language models:

\begin{itemize}
    \item \textbf{Performance of GPT-3.5-based Models:} The GPT-3.5-based models outperform all other models by a wide margin, surpassing even the dedicated Arabic-centric JAIS model's average score by 50\%. This is surprising since the proportion of Arabic language in JAIS's pre-training dataset is around 1000 times larger than in GPT3's pre-training dataset, and JAIS's tokenizer is \textit{a priori} more adapted to Arabic than GPT's (see section \ref{sec:base_models}). 
    
    \item \textbf{Reliability of the Human Score:} The human score serves as the gold standard, highlighting the nuanced comprehension humans have over automated metrics in assessing the quality of legal ruling prediction.
    
    \item \textbf{GPT Score Limitations:} The GPT score, though indicative, showcases its limitations in several instances, rendering it potentially misleading. Moreover, the significant divergence between GPT scores and human evaluations, especially on translated datasets, underscores potential translation errors or inherent metric limitations.
    
    \item \textbf{Inefficiency of ROUGE and BLEU:} The ROUGE and BLEU scores, originally designed for translation and summarization tasks, exhibit their unsuitability for the task at hand.
    
\end{itemize}

\unskip

\unskip







\section{Conclusions}
This study represents a pioneering effort in the realm of Arabic court decision analysis, shedding light on the efficacy of advanced language models in predicting legal outcomes. The findings underscore the remarkable out-performance of GPT-3.5-based models, surpassing even domain-specific models tailored for Arabic language. This unexpected outcome challenges conventional assumptions about the importance of domain specificity and dataset size in model performance. Nevertheless, in spite of the relative superiority of GPT-3.5-based models, their absolute performance on predicting Arabic legal rulings is still unsatisfactory, with an average human score of 2.4 out of 5. Better models fine-tuned on larger Arabic legal datasets need to be developed before LLMs can act as useful legal assistants.

Furthermore, the study emphasizes the indispensable role of human evaluation as the gold standard for assessing the quality of legal ruling predictions. While automated metrics like GPT scores, ROUGE, and BLEU can provide valuable indications in some cases, they exhibit limitations in capturing the nuanced and context-dependent nature of legal language. The inefficacy of ROUGE and BLEU scores in this context underscores the need for tailored evaluation metrics when applying advanced language models to legal analysis tasks. Future research in this domain should focus on developing more contextually relevant evaluation measures to better reflect the accuracy and relevance of predictions in the legal context.

Overall, this study serves as a foundation for future research at the intersection of computational linguistics and Arabic legal analytics. It encourages further exploration into the potential of large language models in assisting legal professionals and policy-makers, ultimately contributing to the effective functioning of the judicial system and the enhancement of legal decision-making processes.


\vspace{6pt} 

\bibliography{biblio}
\end{document}